%% file: paper_arxiv.tex
\documentclass[USenglish,onecolumn]{article}

\usepackage[utf8]{inputenc}
\usepackage{enumerate}
\usepackage{amsfonts}
\usepackage{graphicx}    
\usepackage{algorithmic}
\usepackage{algorithm}
\usepackage{amsthm}
\usepackage{amsmath}
\usepackage{wrapfig}
\usepackage{subfigure}
\usepackage{pdflscape}
\usepackage{ctable}
\usepackage{epstopdf}
\usepackage{lineno}
\usepackage[titletoc,title]{appendix}
\usepackage[symbol]{footmisc}
\usepackage{bm}
\usepackage[round]{natbib}
\usepackage{color}
\usepackage[
hypertexnames=false,
setpagesize=false,
pdfborder={0 0 0},
pdfstartview=Fit,
bookmarksopen=true,
bookmarksnumbered=true
]{hyperref}
\usepackage[hyphenbreaks]{breakurl}

\usepackage{alltt}
\usepackage{framed}

\definecolor{red}{RGB}{255,0,0}
\definecolor{green}{RGB}{0,255,0}
\definecolor{blue}{RGB}{0,0,255}
\definecolor{turquoise}{RGB}{30,144,255}

\theoremstyle{definition}

\providecommand{\keywords}[1]{\textbf{Keywords:} #1}

\newcommand{\argmax}{\operatornamewithlimits{argmax}}

\newcommand{\ceil}[1]{\left\lceil #1 \right\rceil}

\newcommand{\bmhead}[1]{\section*{#1}}

\newcommand{\PROMPT}[1]{\begin{framed}\noindent\textsf{#1}\end{framed}}
\newcommand{\RESP}[1]{\begin{framed}\noindent\textit{#1}\end{framed}}

\newcommand{\RESPP}[1]{
	\begin{framed}
		\makeatletter
		\def\verbatim@font{\normalfont\itshape}
		\makeatother
		\begin{alltt}
			#1
		\end{alltt}
	\end{framed}
}

\begin{document}

\LARGE
\begin{center}
 Pair Programming with Large Language Models for Sampling and Estimation of Copulas

  \vspace{12pt}
  \large

  Jan~Górecki
  \vspace{5pt}

  \small
Department of Informatics and Mathematics, Silesian University in Opava, Univerzitní náměstí 1934/3, 733 40 Karviná, Czech Republic\\
gorecki@opf.slu.cz\\
\end{center}

\normalsize

\renewcommand\thefootnote{\arabic{footnote}}

\begin{abstract}
\input{abstract}
\end{abstract}


\vspace{12pt}


\input{body}

\bibliographystyle{apalike}
\bibliography{lit}

\end{document}

%% file: abstract.tex
Without writing a single line of code by a human, an example Monte Carlo simulation based application for stochastic dependence modeling with copulas is developed using a state-of-the-art large language model (LLM) fine-tuned for conversations. This includes interaction with ChatGPT in natural language and using mathematical formalism, which, under careful supervision by a human-expert, led to producing a working code in MATLAB, Python and \textsf{R} for sampling from a given copula model, evaluation of the model's density, performing maximum likelihood estimation, optimizing the code for parallel computing for CPUs as well as for GPUs, and visualization of the computed results. In contrast to other emerging studies that assess the accuracy of LLMs like ChatGPT on tasks from a selected area, this work rather investigates ways how to achieve a successful solution of a standard statistical task in a collaboration of a human-expert and artificial intelligence (AI). Particularly, through careful prompt engineering, we separate successful solutions generated by ChatGPT from unsuccessful ones, resulting in a comprehensive list of related pros and cons. It is demonstrated that if the typical pitfalls are avoided, we can substantially benefit from collaborating with an AI partner. For example, we show that if ChatGPT is not able to provide a correct solution due to a lack of or incorrect knowledge, the human-expert can feed it with the correct knowledge, e.g., in the form of mathematical theorems and formulas, and make it to apply the gained knowledge in order to provide a solution that is correct. 
Such ability presents an attractive opportunity to achieve a programmed solution even for users with rather limited knowledge of programming techniques.

%% file: body.tex
\section{Introduction}\label{sec1}
The recent progress in solving natural language processing (NLP) tasks using large language models (LLMs) resulted in models with previously unseen quality of text generation and contextual understanding. These models, such as BERT \citep{devlin2018bert}, RoBERTa \citep{liu2019roberta} and GPT-3 \citep{brown2020language}, are capable of performing a wide range of NLP tasks, including text classification, question-answering, text summarization, and more. 
With more than 100 million users registered in two months after release for public testing through a web portal\footnote{\href{https://www.demandsage.com/chatgpt-statistics/}{https://www.demandsage.com/chatgpt-statistics/}}, ChatGPT\footnote{\href{https://openai.com/blog/chatgpt/}{https://openai.com/blog/chatgpt/}} is the LLM that currently most resonates in the artificial intelligent (AI) community. This conversational AI is fine-tuned from the GPT-3.5 series with reinforcement learning from human feedback  \citep{christiano2017deep,stiennon2020learning}, using nearly the same methods as InstructGPT \citep{ouyang2022training}, but with slight differences in the data collection setup. In March 2023, the ChatGPT's developer released a new version of GPT-3.5, GPT-4 \citep{gpt4}. In the time of the writing of this paper, GPT-4 was not freely available, so our results do not include its outputs. However, as a technical report for some of the model's properties is available, we add the relevant information where appropriate.

A particular result of the ChatGPT's fine-tuning is that it can generate corresponding code in many programming languages given a task description in natural language. This can be exploited in \emph{pair programming} \citep{williams2001integrating} with ChatGPT, which then offers several benefits, including:
\begin{itemize}
	\item Enhanced productivity: ChatGPT can help automate repetitive and time-consuming programming tasks, freeing up time for developers to focus on higher-level problem-solving and creative work. On average a time saving of 55\% was reported for the task of writing an HTTP server in JavaScript in the study conducted by the GitHub Next team\footnote{\href{https://github.blog/2022-09-07-research-quantifying-github-copilots-impact-on-developer-productivity-and-happiness/}{https://github.blog/2022-09-07-...}} for GitHub Copilot\footnote{\href{https://github.com/features/copilot}{https://github.com/features/copilot}}. The latter is another code suggestion tool that generates code snippets based on natural language descriptions, powered by  an LLM similar to ChatGPT, Codex \citep{chen2021evaluating}. 
	\item Improved code quality: Pair programming with ChatGPT can help identify errors and bugs in the code before they become bigger problems. ChatGPT can also suggest improvements to code architecture and design.
	\item Knowledge sharing: ChatGPT can help less experienced developers learn from more experienced team members by providing suggestions and guidance.
	\item Better code documentation: ChatGPT can help create more detailed and accurate code documentation by generating comments and annotations based on the code.
	\item Accessibility: ChatGPT can make programming more accessible to people who may not have a programming background, allowing them to collaborate with developers and contribute to projects in a meaningful way. For example, having developed a new theory that requires computations, it might be appealing and time-effective for researchers to use tools like ChatGPT to implement the solution without the need to involve typically expensive manpower in software engineering.
\end{itemize}

Currently, there appear several studies that assess the accuracy of LLMs like ChatGPT based on a set of tasks from a particular area. For example, multiple aspects of mathematical skills of ChatGPT are evaluated in \cite{frieder2023mathematical}, with the main observation that it is not yet ready to deliver high-quality proofs or calculations consistently. In \cite{katz2023gpt},
a preliminary version of GPT-4 was experimentally evaluated against prior generations of GPT on the entire Uniform Bar Examination (UBE)\footnote{\href{https://www.ncbex.org/exams/ube/}{https://www.ncbex.org/exams/ube/}}, and it is reported that GPT-4 significantly outperforms both human test-takers and prior models, demonstrating a 26\% increase over the GPT-3.5-based model and beating humans in five of seven subject areas.
In \cite{bang2023multitask}, an extensive evaluation of ChatGPT using 21 data sets covering 8 different NLP tasks such as summarization, sentiment
analysis and question answering is presented. The authors found that, on the one hand, ChatGPT outperforms LLMs with
so-called zero-shot learning \citep{brown2020language} on most tasks and even out-performs fine-tuned models on some tasks. On the other hand, they conclude that ChatGPT suffers from hallucination problems like other LLMs and it generates more extrinsic hallucinations from its
parametric memory as it does not have access to an external knowledge base. Interestingly, the authors observed in several tasks that the possibility of interaction with ChatGPT enables human
collaboration with the underlying LLM to improve its performance.

The latter observation is the main focus of this work. Rather than evaluating the accuracy of LLMs, we investigate 
ways to benefit from pair programming with an AI partner in order to achieve a successful solution of a task requiring intensive computations. Despite many impressive recent achievements of state-of-the-art LLMs, achieving a functional code is far from being straightforward; one of many unsuccessful attempts is reported at freeCodeCamp\footnote{\href{https://www.freecodecamp.org/news/pair-programming-with-the-chatgpt-ai-how-well-does-gpt-3-5-understand-bash/}{https://www.freecodecamp.org/news/pair-programming...}}. 
Importantly, successful attempts are also emerging. In \cite{maddigan2023chat2vis}, the authors report that LLMs together with the proposed prompts can offer a reliable approach to rendering visualisations from natural language queries, even when queries are highly misspecified and underspecified. However, in many areas, including computationally intensive solutions of analytically intractable statistical problems, a study that demonstrates benefits from pair programming with an AI partner is missing.

This work fills this gap and considers applications involving \emph{copulas} \citep{Nel06,joe2014dependence} as models for stochastic dependence between random variables. These applications are known for their analytical intractability, hence, the Monte Carlo (MC) approach is most widely used to compute the involved quantities of interest. As the MC approach often involves large computation efforts, conducting a MC study requires one to implement all underlying concepts. We demonstrate how to make ChatGPT produce a working implementation for such an application by interacting with it in a natural language and using mathematical formalism. 
To fully illustrate the coding abilities of ChatGPT, the human role is pushed to an extreme, and all the mentioned tasks are implemented without a single line of code written by the human or tweaking the generated code in any way. 
It is important to emphasize that even if the application under consideration relates to a specific area of probability and statistics, our observations apply in a wider scope as the tasks we consider (sampling from a given (copula) model, evaluation of the model's density, performing maximum likelihood estimation, optimizing the code for parallel computing and visualization of the computed results) commonly appear in many statistical applications. Also, we do not present just one way to achieve a successful solution for a given task. Most of the successful solutions are complemented with examples demonstrating which adjustments of our prompts for ChatGPT turn unsuccessful solutions to successful ones. This results in a comprehensive list of related pros and cons, suggesting that if the typical pitfalls are avoided, we can substantially benefit from a collaboration with LLMs like ChatGPT.
Particularly, we demonstrate that if ChatGPT is not able to provide a correct solution due to limitations in its knowledge, it is possible to feed it with the necessary knowledge, and make ChatGPT apply this knowledge to provide a correct solution. 
Having all the sub-tasks of the main task successfully coded in a particular programming language, we also demonstrate how to fully exploit several impressive abilities of ChatGPT. For example, by a simple high-level prompt like ``Now code it in Python.'', ChatGPT correctly transpiles the code from one to another programming language in a few seconds. Also, if an error in the code produced by ChatGPT is encountered during execution, it is demonstrated that ChatGPT is not only able to identify the error, but even immediately produces a corrected version after the error message is copy-pasted to ChatGPT's web interface. 

The paper is organized as follows. Section~\ref{sec:methodology} presents the tasks we consider and sets up the way we interact with ChatGPT. Section~\ref{sec:coding} presents the development of the task via pair programming with ChatGPT. Section~\ref{sec:summary} summarizes the pros and cons observed during the task development, including a discussion on how to mitigate the latter, and Section~\ref{sec:conclusion} concludes.

\section{Methodology} \label{sec:methodology}

\subsection{The task}\label{sec:task}
Let $(x_{ij}) \in \mathbb{R}^{n\times d}$ be a sample of size $n$ from a random vector $(X_1, \dots, X_d) \sim F$, where $F$ is a joint distribution function with the continuous univariate margins $F_1, \dots, F_d$ and \emph{copula} \citep{Skl59} $C$ implicitly given by $F(x_1,\dots, x_d)=C(F_1(x_1),\dots, F_d(x_d))$ for $x_1,\dots, x_d\in \mathbb{R}$. An explicit formula for $C$ is $C(u_1, \dots,$ $u_d) = F(F_1^{-1}(u_1), \dots, F_d^{-1}(u_d)), ~u_1, \dots, u_d \in [0,1]$. A typical application involving the MC approach and copulas assumes that $C$ is unknown but belongs to a parametric family of copula models $\{C_{\theta} : \theta \in \Theta\}$, where $\Theta$ is an open subset of $\mathbb{R}^p$ for some integer $p \geq 1$. The following steps are then considered:
\begin{enumerate}
	\item Estimate the true but unknown parameter $\theta_0 \in \Theta$ of $C_{\theta_0} = C$, e.g., using the pseudo maximum likelihood (ML) estimator
	\begin{equation} \label{eq:MLE}
		\hat{\theta} = \argmax_{\theta \in \Theta}\sum_{i=1}^n \log c_\theta(\hat{u}_{i1}, \dots, \hat{u}_{id}),
	\end{equation}
	where $c_\theta$ is the density of $C_\theta, ~\theta \in \Theta$, $\hat{u}_{ij} = \hat{F}_j(x_{ij})$ and $\hat{F}_j$ is an estimate of $F_j$ for $i \in \{1,\dots, n\}, ~j \in \{1, \dots. d\}$. For some copula families, e.g., for Archimedean ones, evaluation of $c_\theta$ for large $d$ is already a challenge; see \cite{Hofert13}. For pair-copula constructions, the main challenge lies in computing \eqref{eq:MLE}, see \cite{hobaek2013parameter} or \cite{schellhase2018estimating}, typically done using numerical methods like gradient descent. 
	\item Generate a sample $(v_{ij}) \in [0,1]^{N\times d}$ from $C_{\hat{\theta}}$, typically with $N \gg n$. For several popular copula families, this task is also challenging, and involves different techniques for efficiently sampling from $C_{\hat{\theta}}$; see, e.g., \cite{Hof10book,hofert2018hierarchical} for sampling techniques related to Archimedean and {\tiny }Archimax copulas and their hierarchical extensions.
	\item Compute a sample from an analytically intractable distribution, e.g., from the distribution of $\bar{X} = \frac{1}{d}\sum_{j=1}^{d}X_j$. Compared to the previous two points, this is a trivial task, we just need to evaluate $\bar{x}_i = \frac{1}{d}\sum_{i=1}^{d}\hat{F}_j^{-1}(v_{ij}), ~i \in \{1,\dots, N\}$.
	\item Compute the desired quantity based on $\bar{x}_1,\dots,\bar{x}_{N}$. For example, if $X_1, \dots,$ $X_d$ represent risk factor changes, a quantile of the distribution function of $\bar{X}$ represents the \emph{Value-at-Risk} VaR$_\alpha$, commonly used in quantitative risk management; see \cite{mcneil2015quantitative}. Approximating VaR $(X_1 + \dots + X_d)$ is also trivial as it just involves computing the order statistics $\bar{x}_{(1)} \leq \dots \leq \bar{x}_{(N)}$ and then picking out $\bar{x}_{(\ceil{\alpha N})}$, where $\alpha \in [0, 1]$ is a desired confidence level.
	In the same realm, the quantity known as \emph{expected shortfall} involves computing the average of the values $\bar{x}_i$ that are larger than VaR$_\alpha$, so again a computationally trivial task. 
\end{enumerate}
In order to clearly see that the code generated by ChatGPT indeed works as expected without the need of an experienced programmer, we deviate a bit from the above outline, while keeping the non-trivial tasks, i.e., the sampling and estimation. We thus prompt ChatGPT to generate code that does the following: 
\begin{enumerate}
	\item Generate a sample from $C_{\theta_0}$, where $\theta_0 \in \mathbb{R}$.
	\item Based on this sample, compute the ML estimator $\hat{\theta}$ of the true parameter $\theta_0$ using \eqref{eq:MLE}.
\end{enumerate}
Then, we repeat these two steps for several values of $\theta_0$, e.g., linearly spaced on some convenient interval of $\mathbb{R}$. If the plot of the pairs of $(\theta_0, \hat{\theta})$ is close to the identity ($\theta-\theta$) plot, then one has strong evidence of a correct statistical sampling and estimation procedure. Finally, to allow for scaling, we ask ChatGPT to optimize the generated code for parallel computing on CPUs as well as on GPUs.

\subsection{The communication protocol} \label{sec:interaction}
When interacting with ChatGPT, we use the web portal provided by its development team\footnote{\href{chat.openai.com}{chat.openai.com}}. Also, we set up and follow this \emph{communication protocol}:
\begin{enumerate}
	\item We prompt ChatGPT to generate code for solving a selected task in natural language and using mathematical formalism, that is, we specify the task in plain text and do not use any specific formal language. For formulas, we use plain text like \textsf{psi(t) = (1 + t)\^{}(-1/theta)}. 
	\item If the solution generated by ChatGPT is wrong, that is, does not solve the given task, we communicate the problem to ChatGPT, and ask it to provide us with a corrected solution.
	\item If this corrected solution is still wrong, we feed ChatGPT with the knowledge necessary to complete the task successfully, e.g., we provide it with theorems and formulas in plain text. For an example, see the third prompt in Section~\ref{sec:sampling}. 
\end{enumerate}
In this way, we simulate an interaction between two humans, e.g., a client sends by email a task to a software engineer, and we play the role of the client and ChatGPT the role of the software engineer. 
As it is typical that the client is not aware of all details required to solve the task at the beginning of the interaction, such a communication protocol may be frequently observed in practice. The client starts by providing the (subjectively) most important features of the problem in order to minimize her/his initial effort, and then, if necessary, she/he adds more details to get a more precise solution. Importantly, this communication protocol led to a successful completion of the aforementioned tasks, which is reported in Section~\ref{sec:coding}. 

With regards to passing ChatGPT the required knowledge, it is important to realize that ChatGPT does \emph{not} have any memory to remember the previous conversation with a user. Instead, the trick for ChatGPT to appear to remember previous conversations is to feed it the entire conversation history as a single prompt. This means that when a user sends a message, the previous conversation history is appended to the prompt and then fed to ChatGPT.
This prompt engineering technique is widely used in conversational AI systems to improve the model's ability to generate coherent and contextually appropriate responses. However, it is just a trick used to create the \emph{illusion} of memory in ChatGPT. 

If the previous conversation is too long (larger than 4096 tokens\footnote{\href{https://platform.openai.com/docs/models/gpt-3-5}{https://platform.openai.com/docs/models/gpt-3-5}}, where a token is roughly 3/4 of an English word\footnote{The total count of tokens in a piece of text can be precisely measured by \href{https://platform.openai.com/tokenizer}{https://platform.openai.com/tokenizer}}), it may not fit entirely within the context window that ChatGPT uses to generate responses. In such cases, the model may only have access to a partial view of the conversation history, which can result in the model seeming like it has forgotten some parts of the conversation. To mitigate this issue, conversational AI designers often use techniques like truncating or summarizing the conversation history to ensure that it fits within the context window. The way we solve this problem in our example task is re-introducing the parts that we referred to to ChatGPT. For example, when transpiling the code from Python (Appendix~\ref{app:python}) to \textsf{R} (Appendix~\ref{app:R}), we first copy-paste the Python code to ChatGPT's web interface and then ask it to transpile it to \textsf{R}. Without having this technical limitation in mind, it is unlikely to get a correct answer/solution if we refer to the conversion part that does not fit within the context window. Finally note that according to its technical report, GPT-4 uses the context window that is 8x larger than of ChatGPT, so it can contain roughly 25,000 words. This suggests that the limitation imposed by the context window length will become less and less of a concern.

\subsection{The copula family}
In order to make the example task specified in Section~\ref{sec:task} precise, we choose the parametric family $\{C_{\theta} : \theta \in \Theta\}$ to be the popular family of Clayton copulas \citep{clayton_1978}, given by 
\begin{equation}\label{eq:clayton_gen}
	C_\theta(u_1,u_2) = (\max(u_1^{-\theta} + u_2^{-\theta} - 1, 0))^{-\frac{1}{\theta}}, 
\end{equation}
where $u_1,u_2\in[0,1]$ and $-1\leq\theta<\infty,\,\theta\neq 0$. 
This family of copulas is used in a wide variety of applications. To mention several recent ones, e.g., \cite{huang2022spatio} use it to analyse the correlation between the residual series of a long short-term memory neural network and a wind-speed series. Particularly, the maximum likelihood estimation of the copula parameter is utilized, i.e., the procedure that ChatGPT implements here in Section~\ref{sec:estimation}. In the simulation study in \cite{michimae2022likelihood}, where copula-based competing risks models for latent failure times are proposed, the authors utilize sampling from Clayton copulas, i.e., the procedure that ChatGPT implements here in Section~\ref{sec:sampling}.

For simplicity, as well as the fact that the models with $\theta < 0$, that is, those with negative dependence,  are rarely used in practice, we restrict to $\theta \in (0, \infty)$, which allows one to rewrite  \eqref{eq:clayton_gen} to 
\begin{equation}\label{eq:clayton}
	C_\theta(u_1,u_2) = (u_1^{-\theta} + u_2^{-\theta} - 1)^{-\frac{1}{\theta}}. 
\end{equation}
The technical reason for choosing this family is its simple analytical form, which makes easier for the reader to track all the formulas we ask for and get from ChatGPT, e.g., the probability density function (PDF). Another reason is ChatGPT's relatively limited knowledge of this family. By contrast, e.g., for the most popular family of Gaussian copulas, ChatGPT was not able to generate a sampling algorithm without being fed with some necessary theory. The latter simulates a realistic situation when ChatGPT is facing a new theory/concept, e.g., one recently developed by the user. However, we would like to encourage the reader to experiment with any family of interest or even with a task that differs from our example. 

\section{Pair programming with ChatGPT}\label{sec:coding}

Pair programming \citep{williams2001integrating} is a software development technique in which two programmers work together at one workstation. One, the \emph{driver}, writes code while the other, the \emph{navigator} reviews each line of code as it is typed in and considers the ``strategic'' direction of the work. In what follows, we, the human-part of the pair, put ourselves in the role of the navigator, giving specific tasks to the driver, ChatGPT, representing the AI-part of the pair.

We begin prompting ChatGPT with tasks that require rather less lines of code in order to not overwhelm the reader with the amount of results, while still demonstrating as many abilities of ChatGPT as possible. Then, we gradually move to larger tasks until we get all of them implemented. Our prompts are in \textsf{sans serif}. Responses of ChatGPT are either in \texttt{verbatim}, if the response is code, or in \textit{italic}, otherwise. In its first occurrence, we indicate this explicitly. Note that as ChatGPT is quite a loquacious LLM, we mostly limit the length of its responses, otherwise we would be overwhelmed with answers of unnecessary detail. For the same reason, when asking ChatGPT for code, we mostly omit its comments on the produced code. Also note that when commenting on ChatGPT's responses, we speak of it as if it were a human, e.g., it ``understands'', ``knows'' or ``is aware of'' something, which should be interpreted by the reader in the sense that ChatGPT produced a response that (typically very well) \emph{mimics} a corresponding human reaction.
As ChatGPT's responses are by default non-deterministic, i.e., giving it the same prompt again, the response might slightly differ. To take this feature into account, we re-generate the response for each of our prompts three times, and if these responses are factually different from each other, we indicate it accordingly. Finally note that the whole interaction is conducted in one chat window. Once we observe that ChatGPT starts to forget the previous context due to the reasons described in Section~\ref{sec:interaction}, we re-introduce it as we describe in the same section.

In the rest of this section, we first investigate the knowledge of ChatGPT on the topic under consideration, and then we prompt it to generate code for evaluation of the density of the Clayton copula, for ML estimation of its parameter, for sampling from the copula, for creating a visualization of the example Monte Carlo approach, and for optimizing the code for parallel computations. 

\subsection{Warm up}
\label{sec:warm_up}


\PROMPT{(Prompt:) What is the Clayton bivariate copula, why is it interesting, which is its analytical form and is it an Archimedean copula? Respond to these questions in at most 100 words.}
\begin{framed}
\makeatletter
\def\verbatim@font{\normalfont\itshape}
\makeatother
\vspace{-10pt} \begin{alltt}
(Response:) The Clayton copula is a bivariate copula function that is widely used in statistics and finance to model dependence between variables. It is interesting because it has a simple form and can capture positive dependence between variables while allowing for asymmetric tails. The analytical form of the Clayton copula is C(u, v; theta) = (max(u^(-theta) + v^(-theta) - 1, 0))^(-1/theta), where theta is the Clayton copula parameter. The Clayton copula is an Archimedean copula, meaning that it can be expressed as the generator function of a particular class of symmetric distributions.
\end{alltt} \vspace{-10pt} 
\end{framed}

We see that ChatGPT can save our time by quickly and concisely summarizing basic facts about the topic of our interest. We can also limit the size of the answer, which is satisfied in this 91 words long answer. The information about the positive dependence probably follows from what we have already stated before: the negatively dependent models are rarely used in practice, which is probably reflected in ChatGPT's training data. 
However, several details of the answer can be discussed. In lines 3 and 4, ``random variables'' instead of just ``variables'' would be more precise. From the last sentence, it follows that an Archimedean copula can be expressed as the generator function for some symmetric distributions. This is at least confusing as Archimedean copulas are rather a particular class of copulas admitting a certain functional form \emph{based} on so-called generator functions. Finally, \emph{symmetric} distributions have their precise meaning: such a distribution is unchanged when, in the continuous case, its probability density function is reflected around a vertical line at some value of the random variable represented by the distribution. Whereas Archimedean copulas posses a kind of symmetry following from their exchangeability, they do not belong to symmetric distributions.

To investigate the limits of ChatGPT's knowledge, let us prompt it with two further questions. According to the previous response, we can speculate that it has limited knowledge on the Clayton models with negative dependence. 
\PROMPT{If the Clayton copula parameter is negative, e.g., theta = -1/2, is the corresponding Clayton copula a model for positive dependence between variables? Respond with one word.}
\RESP{No.}
\RESP{Correction: Yes.}
\RESP{Yes.}
\noindent We prompted ChatGPT to answer the same question three times and got contradicting answers. The first answer is correct, however, after asking again, ChatGPT changed its mind. Before commenting on that, let us try once again, with a bit more complex concept.
\PROMPT{If the Clayton copula parameter is positive, e.g., theta = 1, is the survival copula of the corresponding Clayton copula a model with lower or with upper tail dependence?}
\RESP{When the Clayton copula parameter is positive, i.e., theta $>$ 0, the survival copula of the corresponding Clayton copula is a model with lower tail dependence.}
\noindent If $(U_1, U_2)\sim C$, then the survival copula of $C$ is the distribution of $(1-U_1, 1-U_2)$, and thus the properties of the lower tail of $C$ are the properties of the upper tail of the survival copula. Hence, we got an incorrect answer. After asking again, we got this response. 
\RESP{For the Clayton copula, if the parameter theta is positive, the survival copula will have upper tail dependence. This is because the Clayton copula exhibits positive lower tail dependence, which means that when one variable has a very low value, the other variable tends to have a low value as well. This implies that when one variable has a very high survival time, the other variable also tends to have a high survival time, leading to upper tail dependence in the survival copula.}
\noindent Again, we got contradicting answers. Based on this observation, the reader could raise the following question.
\PROMPT{Would you agree with the following statement? ChatGPT is a large language model, and thus, if we drastically simplify it, it basically makes inference on which word comes next. However, it makes no inference on the underlying concepts.}
\RESP{As an AI language model, ChatGPT is indeed designed to make predictions on the next word based on the context and patterns in the input text. However, it is also capable of capturing and inferring underlying concepts from the text, to some extent. While it is true that ChatGPT is not capable of understanding concepts in the same way as humans do, it can still recognize patterns and relationships between words and phrases, and use this information to make inferences and generate responses that reflect some level of understanding of the underlying concepts.}
\noindent This response confirms what we have seen so far, hence, any user should take these limitations into account with the utmost seriousness and be extremely careful when asking ChatGPT for some reasoning. The examples above also well illustrate that the current version of ChatGPT is definitely \emph{not} an appropriate tool for reasoning, which is as also observed by \cite{frieder2023mathematical} and \cite{bang2023multitask}.
However, this by no means implies that it cannot serve as a helpful AI partner for pair programming.

\subsection{The density}\label{sec:density}
It can be easily shown that the density $c_{\theta}$ of $C_{\theta}$ is 
\begin{equation} \label{eq:density}
c_{\theta}(u,v) = \frac{\partial^2C_{\theta}(u,v)}{\partial u \partial v} = (\theta + 1)(uv)^{-(\theta+1)} (u^{-\theta} + v^{-\theta} - 1)^{-\frac{1}{\theta} - 2} 
\end{equation}
for $\theta > 0$.
Before we ask ChatGPT to generate code evaluating the Clayton copula density, it would be good to ask for a plain formula.
\PROMPT{What is the analytical form of the probability density function of the bivariate Clayton copula? Output the formula using LaTeX.}
\begin{framed}
\noindent $$c_{\theta}(u,v) = \frac{\partial^2C_{\theta}(u,v)}{\partial u \partial v} = \frac{\theta + 1}{\theta}\biggl(u^{-\theta-1}v^{-\theta-1} (u^{-\theta} + v^{-\theta} - 1)^{-\frac{\theta+2}{\theta+1}}\biggr).$$
\end{framed}
\noindent Out of many factually distinct responses, we present this one to illustrate how dangerous it could be to believe that ChatGPT knows or can derive the right formula. Even if it looks quite similar to \eqref{eq:density}, this is \emph{not} the density of $C_{\theta}$.
\PROMPT{This formula is not correct. Which is the analytical form of the probability density function of the bivariate Clayton copula? Output the formula using LaTeX.}
\begin{framed}
$$c(u,v) = \left[(u^{-\theta} + v^{-\theta} - 1)^{-\frac{1}{\theta} - 2} \right] \cdot \left(u^{-\theta} v^{-\theta} \right)^{-\frac{1}{\theta} - 1}$$
\end{framed}
\noindent Another formula quite similar to \eqref{eq:density}, but also this one is incorrect. As already mentioned before, ChatGPT is not a good option when it comes to reasoning. So, to get the right formula, some symbolic tool is definitely preferred. However, note that ChatGPT plugins\footnote{\href{https://openai.com/blog/chatgpt-plugins}{https://openai.com/blog/chatgpt-plugins}} has been recently announced, and particularly Code interpreter, which is an experimental ChatGPT model that can use Python, handle uploads and downloads, allows for symbolic computations. Even if not freely available yet, this might be also a possible way to mitigate the problem, directly in the ChatGPT environment.

Following our communication protocol, let us feed ChatGPT with the right formula, and ask for a corresponding function in three programming languages: 1) MATLAB\footnote{We used version R2020a.}, which represents a proprietary software, 2) Python\footnote{Version 3.9.}, an open-source software popular in the AI community, and \textsf{R}\footnote{Version 4.2.2.}, an open-source software popular in the statistical community. 
Note that in cases when the output is too wide, we adjust it in order to fit on the page; otherwise we do not adjust it in any other way.
\PROMPT{This is the correct formula: c(u,v) = (1 + theta) * (u * v)\^{}(-theta-1) * (u\^{}(-theta) + v\^{}(-theta) - 1)\^{}(-2-1/theta). Code it in MATLAB as ClaytonCopulaDensity(u, v, theta).}
\begin{framed}
	\makeatletter
	\def\verbatim@font{\normalfont\itshape}
	\makeatother
	\vspace{-10pt} \begin{alltt}
Here's the MATLAB code for the probability density function of the Clayton copula with parameter theta evaluated at (u, v):
	\end{alltt} \vspace{-10pt}
	\makeatletter
\def\verbatim@font{\normalfont\ttfamily}
	\makeatother
\vspace{-10pt} \begin{verbatim}
function dens = ClaytonCopulaDensity(u, v, theta)
dens = (1 + theta) * (u * v)^(-theta - 1) * ...
       (u^(-theta) + v^(-theta) - 1)^(-2 - 1/theta);
end
\end{verbatim} \vspace{-10pt}
	\makeatletter
\def\verbatim@font{\normalfont\itshape}
\makeatother
\vspace{-10pt} \begin{alltt}
You can call this function with your desired values of u, v, and theta to evaluate the density of the Clayton copula at those values.
\end{alltt} \vspace{-10pt}
\end{framed}
\noindent After feeding ChatGPT with the right formula, it immediately generated a functional code. Notice that we used quite a natural and relaxed form of conversation, e.g., like in an email.
\PROMPT{Code ClaytonCopulaDensity in Python.}
\begin{framed}
\vspace{-10pt} \begin{verbatim}
def ClaytonCopulaDensity(u, v, theta):
    density = ((1 + theta) * (u * v)**(-theta - 1) * 
               (u**(-theta) + v**(-theta) - 1)**(-2 - 1/theta))
    return density
\end{verbatim} \vspace{-10pt}
\end{framed}
\PROMPT{And in R.}
\begin{framed}
	\vspace{-10pt} \begin{verbatim}
ClaytonCopulaDensity <- function(u, v, theta) {
    density <- ((1 + theta) * (u * v)^(-theta - 1) * 
                (u^(-theta) + v^(-theta) - 1)^(-2 - 1/theta))
    return(density)
}
	\end{verbatim} \vspace{-10pt}
\end{framed}
\noindent As ChatGPT takes into account the previous conversation, we could afford to be extremely concise with our prompts and still get correct solutions. In what follows, we ask for code only in MATLAB to save space. However, the equivalent code in Python and \textsf{R} is shown in the appendices, where all the functions can be easily identified by their names.

\subsection{The estimation} \label{sec:estimation}
\PROMPT{Having this ClaytonCopulaDensity(u, v, theta) function, could you write a MATLAB code for the maximum likelihoood estimator of the parameter theta?}
\begin{framed}
\vspace{-10pt} \begin{verbatim}
function thetaMLE = ClaytonCopulaMLE(u, v)
loglik = @(theta) sum(log(ClaytonCopulaDensity(u, v, theta)));
thetaMLE = fminsearch(@(theta) -loglik(theta), 1);
end
\end{verbatim} \vspace{-10pt}
\end{framed}
By contrast to our struggles with the PDF, we immediately got a correct solution. This may be due to the fact that code snippets computing ML estimators occur more frequently in ChatGPT's training data. This pattern (the more general the task, the more frequently we receive a working solution on first trial) is observed also in other examples throughout this work. 

\subsection{The sampling}\label{sec:sampling}
\PROMPT{In MATLAB, code a function ClaytonSample(theta, n) that generates a sample of size n from the bivariate Clayton copula with parameter theta.}

\begin{figure}[htbp]
	\centering
	\includegraphics[width=.325\textwidth]{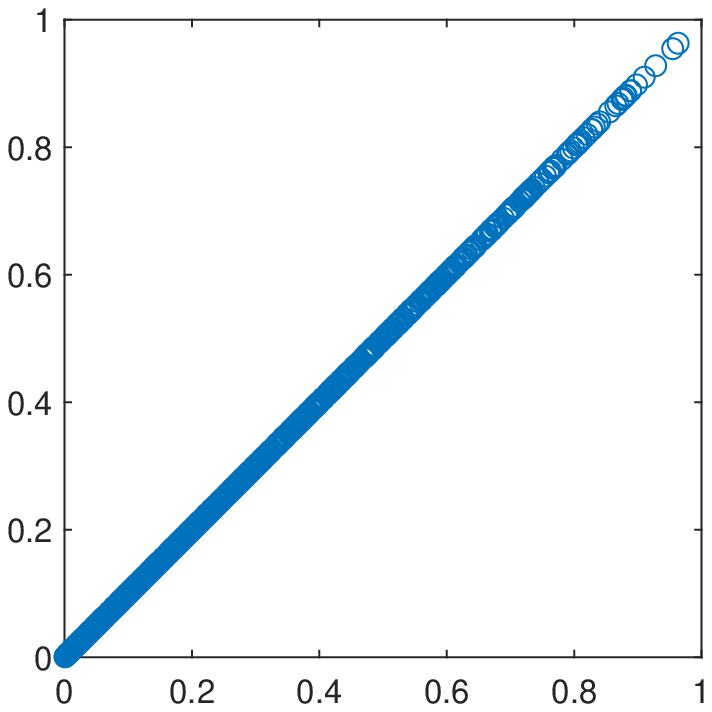}
	\includegraphics[width=.325\textwidth]{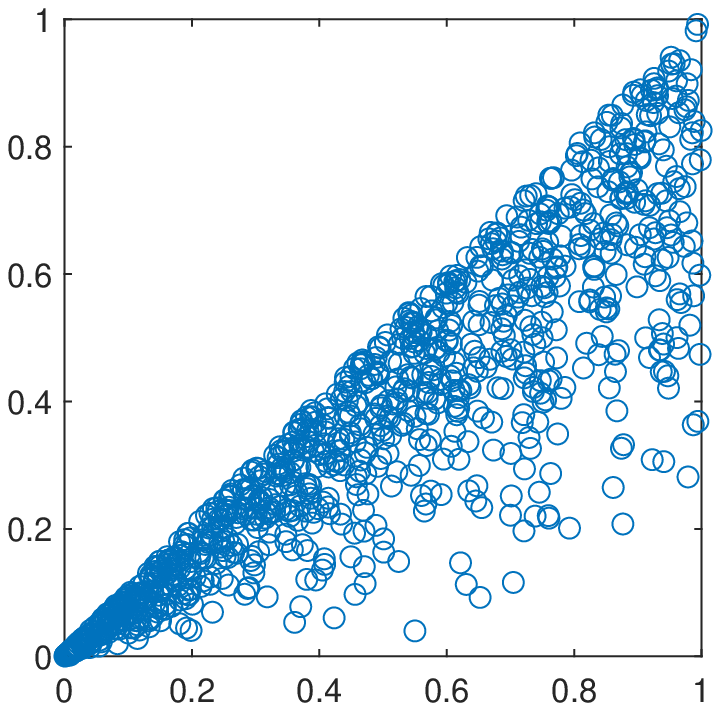}
	\includegraphics[width=.325\textwidth]{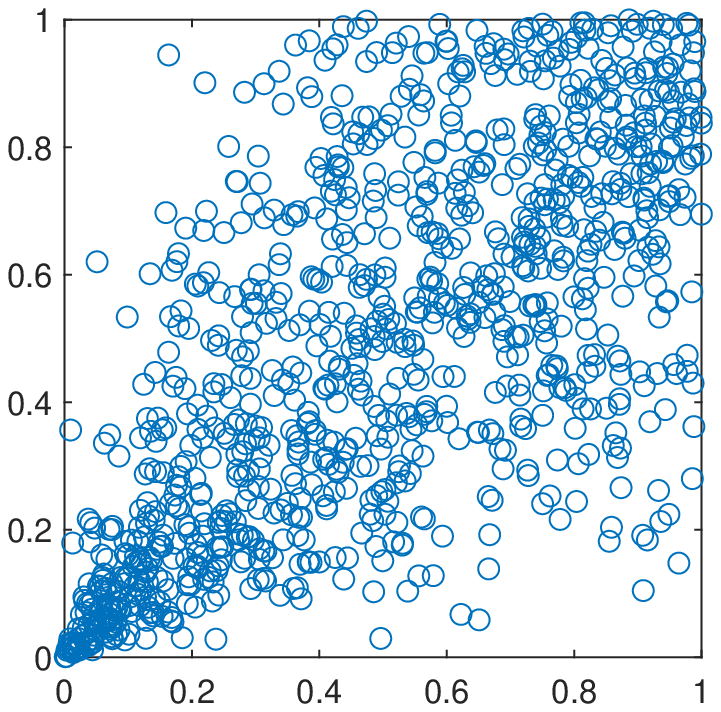}
	\caption{A sample from the random vector $(C(U_1,U_2), C(U_1,U_2))$ (left). A sample from another incorrect sampling trial of ChatGPT (center). A sample generated by the correct sampling procedure based on theory fed to ChatGPT (right). The sample was obtained by calling \texttt{ClaytonSample(2, 1000)}.}
	\label{fig:samples}
\end{figure}

\begin{framed}
\vspace{-10pt} \begin{verbatim}
function u = ClaytonSample(theta, n)
u1 = rand(n, 1);
u2 = rand(n, 1);
v1 = (u1.^(-theta)+u2.^(-theta)-1).^(-1/theta);
v2 = (u1.^(-theta)+u2.^(-theta)-1).^(-1/theta);
u = [v1, v2];
end
\end{verbatim} \vspace{-10pt}
\end{framed}
\noindent We see that \texttt{u1} and \texttt{u2} are drawn from the standard uniform distribution, which is typically used in such sampling algorithms. However, then these are just used as arguments of \eqref{eq:clayton}, resulting in two identical one-dimensional samples from the distribution of $C(U_1,U_2)$, where $U_1$ and $U_2$ are two independent random variables with the standard uniform distribution. Note if the random vector $(U_1,U_2)$ would be distributed according to $C$, then the distribution function of $C(U_1,U_2)$ would be the distribution known as Kendall function; see \citet[pp. 419--422]{joe2014dependence}. So we are witnessing a mixture of approaches related to copula modelling, which, however, do not lead to a correct solution.
A sample generated with this code for $\theta = 2$ is shown at the left-hand side of Figure~\ref{fig:samples}, and it is clearly \emph{not} a sample from the Clayton copula $C_{\theta}$ with parameter $\theta = 2$. 
\PROMPT{In ClaytonSample(theta, n), v1 and v2 are two identical vectors. Hence, u = [v1, v2] cannot be a sample from the bivariate Clayton copula with parameter theta. Could you code a correct version of ClaytonSample(theta, n)?}
\begin{framed}
\vspace{-10pt} \begin{verbatim}
function u = ClaytonSample(theta, n)
w = rand(n, 1);
v = rand(n, 1);
for i = 1:n
  v(i) = (w(i)^(-theta) * (1 - v(i))^ ...
          (-theta/(1+theta)))^(-1/theta);
end
u = [w, v]
\end{verbatim} \vspace{-10pt}
\end{framed}
\noindent In the second try, ChatGPT also failed to produce a correct solution; see the sample generated by this code at the center of Figure~\ref{fig:samples}. Here, the reasoning behind is even less clear. These two trivial examples again illustrate that we must be extremely careful about results produced by ChatGPT. On the other hand, this gives us a chance to show that ChatGPT can interactively learn new concepts by feeding it with the necessary theory as we will demonstrate next.

A standard approach for efficient sampling from Archimedean copulas was introduced in \cite{marshall1988families}. Let us feed ChatGPT with the related theory and ask it for a correct solution. 
\PROMPT{Let psi(t) = (1 + t)\^{}(-1/theta) be a function with parameter theta in (0, infty). Let E1 and E2 be two independent samples from the standard exponential distribution, and V be an independent sample from the Gamma distribution with the shape parameter 1/theta and the scale parameter 1. Then (psi(E1/V), psi(E2/V)) is a sample from the bivariate Clayton copula with parameter theta. Code a MATLAB function ClaytonSample(theta, n), which generates n samples from the bivariate Clayton copula with parameter theta using the provided theory.}

\begin{framed}
	\vspace{-10pt} \begin{verbatim}
function u = ClaytonSample(theta, n)
% Generate two samples from exponential distribution
E = exprnd(1, 2, n);        
% Generate a sample from gamma distribution
V = gamrnd(1/theta, 1, 1, n); 
% Define the function psi(t)
psi = @(t) (1 + t).^(-1/theta); 
% Apply the transformation 
u = [psi(E(1,:) ./ V); psi(E(2,:) ./ V)]'; 
end
	\end{verbatim} \vspace{-10pt}
\end{framed}
\noindent Apart from producing a correct solution, which generated the sample at the right-hand side of Figure~\ref{fig:samples}, this is a clear demonstration of the ability of ChatGPT to \emph{learn} during the interaction with the user (having in mind that this capacity is just a quite convincing illusion enabled by the prompt engineering technique applied to ChatGPT, as discussed in Section~\ref{sec:interaction}). In contrast to the previous example, where it ``only'' translated the fed formula for the PDF to a more-or-less similarly looking code, this example shows that ChatGPT is able to understand even relatively complex concepts. 

Such an ability makes ChatGPT a feasible tool also in cases when it faces unknown concepts. This is essential, e.g., in cases when a new theory developed by a researcher is about to be coded. Also, notice that ChatGPT saves us time by mapping our concepts, e.g., the standard exponential and gamma distribution, to existing functions in the considered programming language. Particularly, without explicitly asking, 
ChatGPT avoided a loop iterating over 1 to \textsf{n} in which one sample from the copula would be generated, which is typically slow, but generates directly \textsf{n} samples of \textsf{E1} and \textsf{E2} from the standard exponential distribution (\texttt{exprnd(1, 2, n)}, where \texttt{1} denotes the parameter of the exponential distribution).
We can thus avoid probably the most boring part of coding when we are browsing the documentation for a particular function in available libraries. 

Finally, let us perform a simple check of all the previously generated functions. 
\PROMPT{Generate a script that draws a sample of size 1000 from the Clayton copula with theta = 2 and then computes an estimate theta\_hat of theta using the ML estimator.}
\begin{framed}
	\vspace{-10pt} \begin{verbatim}
U = ClaytonSample(2, 1000);
theta_hat = ClaytonCopulaMLE(U(:,1), U(:,2));
	\end{verbatim} \vspace{-10pt}
\end{framed}
\noindent Being in the context of the previous conversation, ChatGPT exploited the \texttt{ClaytonCopulaMLE} function generated in Section~\ref{sec:estimation}.
After executing the simple check code, we got the following error.
\begin{framed}
\vspace{-10pt} \begin{verbatim}
Error using  * 
Incorrect dimensions for matrix multiplication. Check that 
the number of columns in the first matrix matches the 
number of rows in the second matrix. To
perform elementwise multiplication, use '.*'.

Error in ClaytonCopulaDensity (line 2)
dens = (1 + theta) * (u * v)^(-theta - 1) * ...

Error in ClaytonCopulaMLE>@(theta)sum(log(...
ClaytonCopulaDensity(u,v,theta))) (line 2)
loglik = @(theta) sum(log(ClaytonCopulaDensity(u, v, theta)));

Error in ClaytonCopulaMLE>@(theta)-loglik(theta) (line 3)
thetaMLE = fminsearch(@(theta) -loglik(theta), 1);

Error in fminsearch (line 200)
fv(:,1) = funfcn(x,varargin{:});

Error in ClaytonCopulaMLE (line 3)
thetaMLE = fminsearch(@(theta) -loglik(theta), 1);
\end{verbatim} \vspace{-10pt}
\end{framed}
\noindent We copy-pasted the error message to ChatGPT, e.g., gave it the prompt
\PROMPT{We got this error message: ***}
\noindent where \textsf{***} was the error message, and got the following response.

\begin{framed}
	\makeatletter
	\def\verbatim@font{\normalfont\itshape}
	\makeatother
	\vspace{-10pt} \begin{alltt}
The error message suggests that the code is trying to perform matrix multiplication with the * operator, but the dimensions of the matrices do not match. This is likely happening in the line dens = (1 + theta) * (u * v)\^{}(-theta - 1) * ....

To fix this, we need to use element-wise multiplication with the .* operator instead of matrix multiplication with *. Here is the corrected version of the function:
	\end{alltt} \vspace{-10pt}
	\makeatletter
	\def\verbatim@font{\normalfont\ttfamily}
	\makeatother
\vspace{-10pt} \begin{verbatim}
function dens = ClaytonCopulaDensity(u, v, theta)
dens = (1 + theta) * (u .* v).^(-theta - 1) .* ...
(u.^(-theta) + v.^(-theta) - 1).^(-2 - 1/theta);
end
\end{verbatim} \vspace{-10pt}
	\makeatletter
\def\verbatim@font{\normalfont\itshape}
\makeatother
\vspace{-10pt} \begin{alltt}
Note the use of .* instead of * and .^ instead of ^ to indicate element-wise operations.
\end{alltt} \vspace{-10pt}
\end{framed}
\noindent ChatGPT not only detected what is wrong, but provided a corrected solution together with an explanation of the problem. Given that these trivial mistakes occur quite often in daily practice, such help that points out to the problem or even solves it can significantly save time.
After executing the simple check with the corrected version of \texttt{ClaytonCopulaDensity}, we got \texttt{theta\_hat = 2.12}. So far so good.

\subsection{The visualization} \label{sec:visualization}
\begin{figure}[t!]
	\centering
	\includegraphics[width=.49\textwidth]{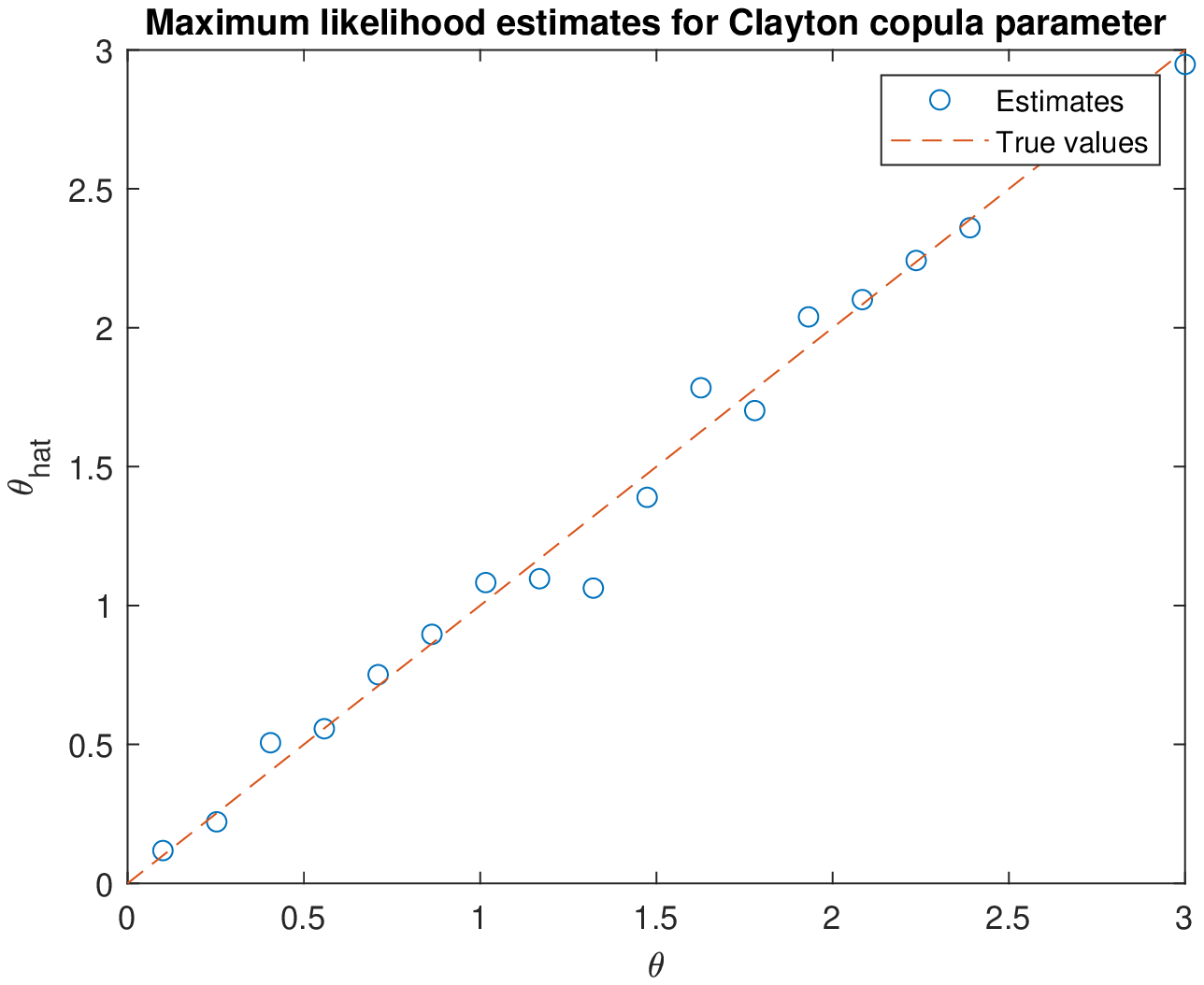}
	\includegraphics[width=.49\textwidth]{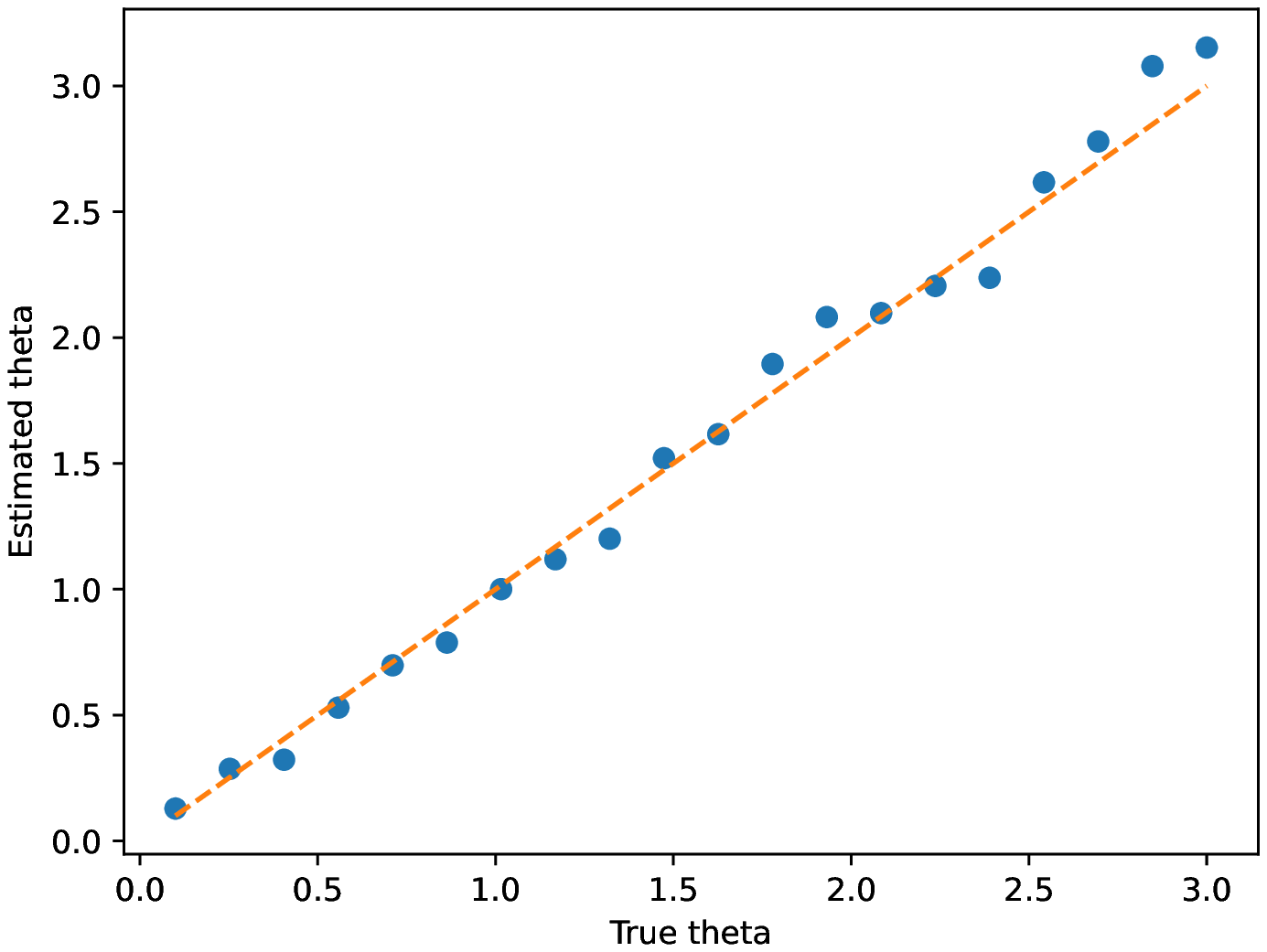}
	\includegraphics[width=.49\textwidth]{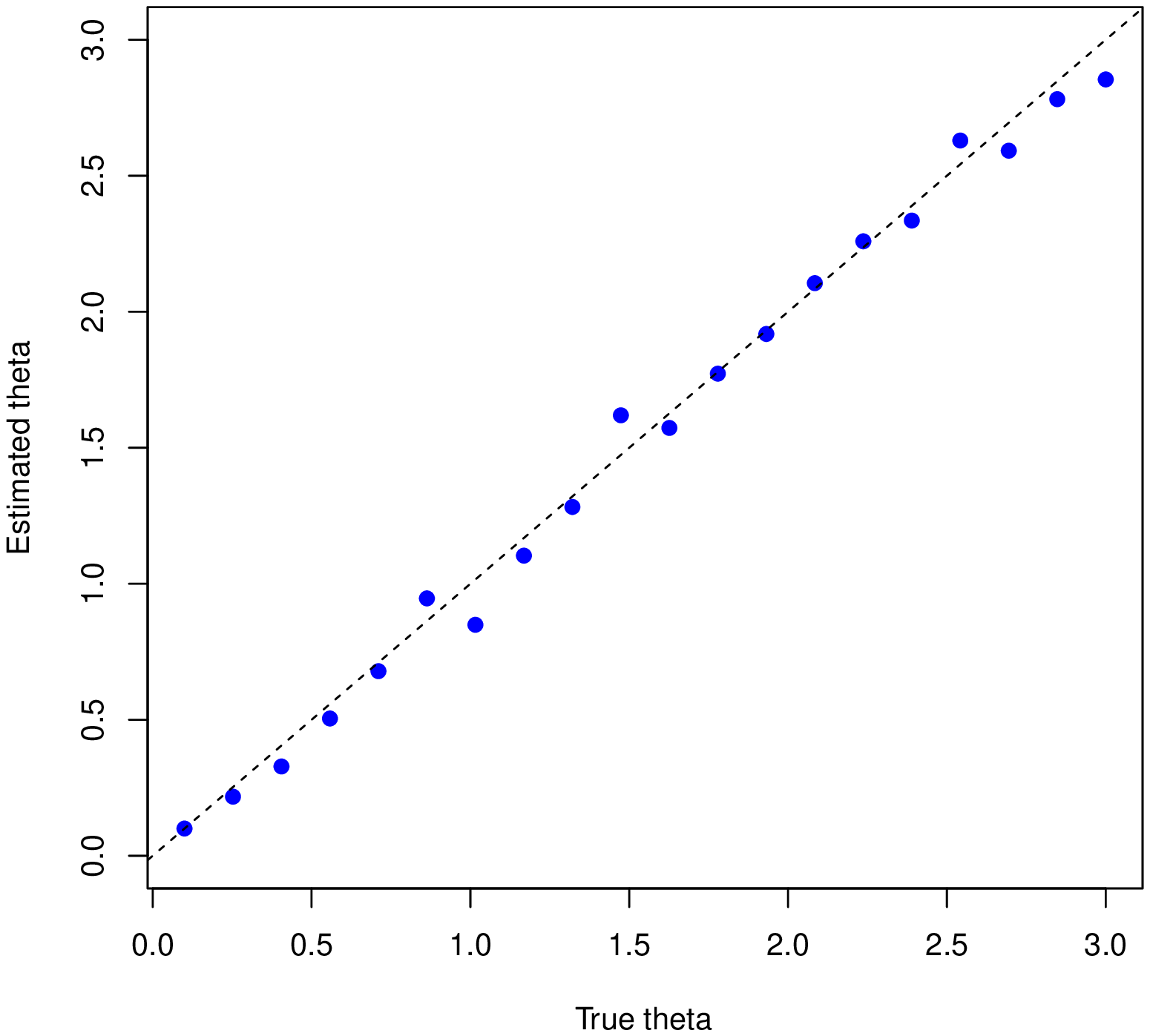}
	\caption{The plots produced by the code generated by ChatGPT in MATLAB (top-left), Python (top-right) and \textsf{R} (bottom).}
	\label{fig:thetas_plot}
\end{figure}

\PROMPT{Write MATLAB code that computes and creates a plot of 20 pairs of (theta, theta\_hat), where theta is from (0.1, 3) and theta\_hat is the ML estimate of theta (computed using ClaytonCopulaMLE(u, v)) based on a sample of size 1000 from the Clayton bivariate copula with parameter theta (generated using U = ClaytonSample(theta, n))}
\begin{framed}
	\vspace{-10pt} \begin{verbatim}
n = 1000;
thetas = linspace(0.1, 3, 20);
theta_hats = zeros(size(thetas));

for i = 1:length(thetas)
  U = ClaytonSample(thetas(i), n);
  theta_hats(i) = ClaytonCopulaMLE(U(:,1), U(:,2));
end

plot(thetas, theta_hats, 'o');
hold on;
plot([0, 3], [0, 3], '--');
xlim([0, 3]);
ylim([0, 3]);
xlabel('\theta');
ylabel('\theta_{hat}');
title(['Maximum likelihood estimates for ' ...
        'Clayton copula parameter']);
legend('Estimates', 'True values');
	\end{verbatim} \vspace{-10pt}
\end{framed}
\noindent The plot generated by the response is depicted on the left-hand side of Figure~\ref{fig:thetas_plot}. 
As the $(\theta, \hat{\theta})$ pairs are close to the identity, it gives an evidence that all previously generated code works properly. We would like to highlight that even if ChatGPT is not instructed to:
\begin{enumerate}
	\item The parameters $\theta$ are \emph{linearly} spaced in the desired interval. This is a typical choice for many visualizations.
	\item It shows perfect identity by a line, also typically considered in an ideal benchmark. This clearly demonstrates that ChatGPT at least partially understands the underlying concepts, i.e., that we are estimating the true value of some parameter. 
	\item Typical time-consuming trivialities like limits, labels, title and legend are also shown.
\end{enumerate}
All in all, this is the type of tasks where the user can substantially benefit from a collaboration with tools like ChatGPT. 

However, we would like to note that two iterations of our last prompt were done before we got the presented one. In the first one, we omitted the content of the two parentheses with the function names. In that case, the output of \texttt{ClaytonSample} did not match the dimension of the input of \texttt{ClaytonCopulaMLE}. In the second iteration, we added those parentheses to the prompt, but without ``\textsf{U = }'', and got a very similar error. We copy-pasted the error message to ChatGPT, but this time it was not succesful to provide a correct version. Finally, we added that ``\textsf{U = }'' in the prompt with the intuition that ChatGPT is not aware of the right dimensions of the output, and this way we got a working solution. The main take-away from this example is that obtaining a correct solution is an iterative process requiring careful prompt engineering. The connection to human-human collaboration is clear: until the software engineer does not understand what the researcher exactly wants, she/he will probably deliver unsatisfactory solutions.

\PROMPT{Code the last functionality from MATLAB in Python.}
\noindent The response is shown in Appendix~\ref{app:python} and the corresponding plot is depicted at right-hand side of Figure~\ref{fig:thetas_plot}. When a project is migrated from one programming language to another, this ability of ChatGPT could limit the related efforts/expenses dramatically.
\PROMPT{And in R.} 
\noindent In this case, the transpilation to \textsf{R} was not successful as we again got a problem with the dimensions matching of the input of \texttt{ClaytonCopulaMLE} and output of \texttt{ClaytonSample}. We were also unsuccessful with copy-pasting the error message to ChatGPT. Clearly, we could be more precise by specifying these dimensions. However, we were deliberately underspecifying the task in order to illustrate what separates successful solutions from unsuccessful ones.

To resolve the task, we finally used a rather ``brute-force'' approach  consisting of copy-pasting the whole code in Python to ChatGPT and asking it  for the same code on \textsf{R}.
The response is shown in Appendix~\ref{app:R} and the corresponding plot is depicted at the bottom of Figure~\ref{fig:thetas_plot}. Another way we obtained a working solution was to feed ChatGPT with explicit information about the dimensions of the inputs and outputs, i.e., using a similar approach as before by adding ``\textsf{U = }''. On the one hand, this approach is more elegant then the brute-force one, on the other hand, the brute-force approach allowed us to get a working solution in less time. A choice between them thus depends on the user's priorities.

Basically we now have an implementation of our example task in every language supported by ChatGPT. Note that an implementation of sampling from the Clayton copula in Python appeared relatively recently as a serious effort presented in \cite{boulin2022sample}. Here, we got an equivalent implementation completely created by an AI. Clearly, with the abilities of ChatGPT, achieving a solution in distinct  programming languages will become much less valuable.

\subsection{The parallelization} \label{sec:parallelization}
Note that as the codes for the following tasks in Python and \textsf{R} were rather long, we present only the MATLAB versions. 
\subsubsection{CPUs} \label{sec:cpus}
\begin{figure}[htbp]
	\centering
	\includegraphics[width=.49\textwidth]{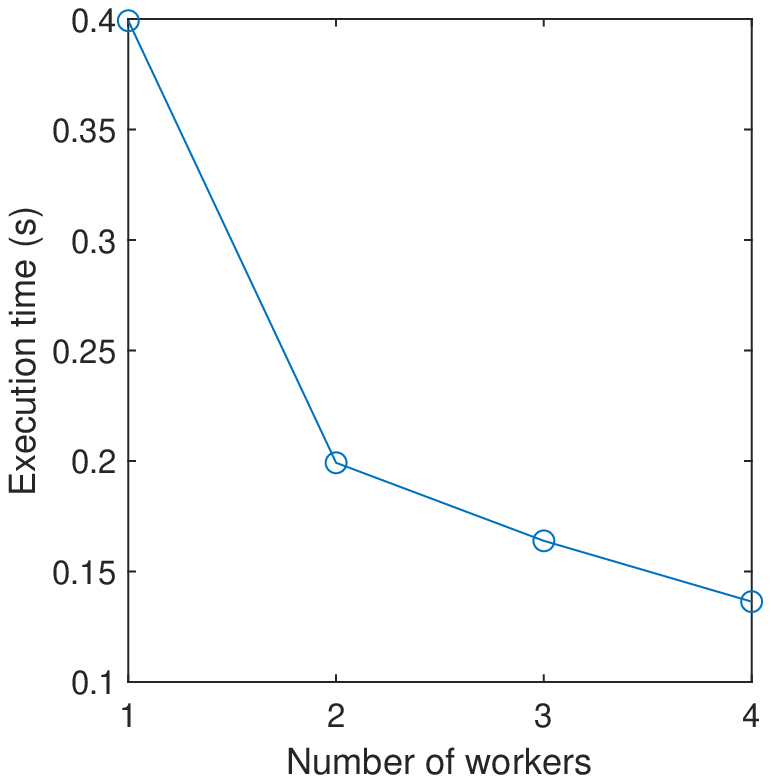}
	\includegraphics[width=.49\textwidth]{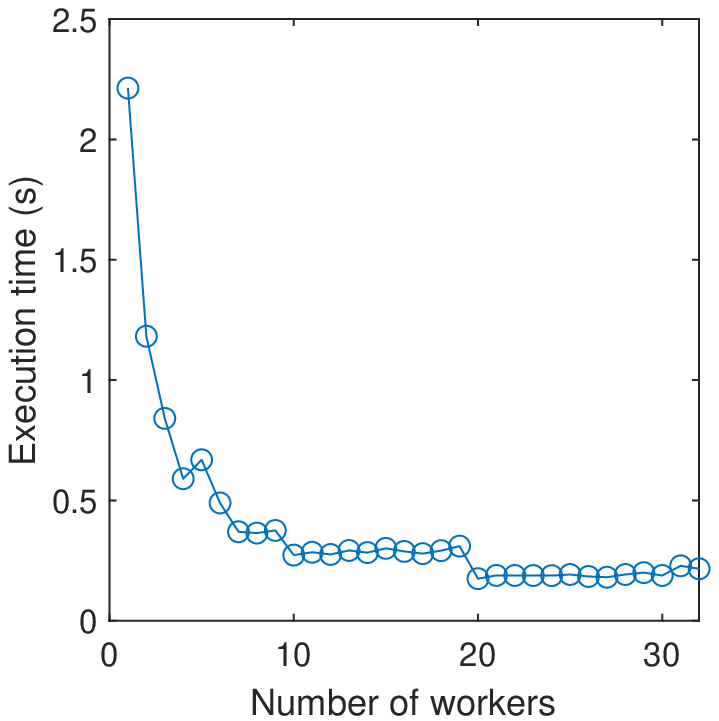}
	\caption{Two plots demonstrating the improvement in run-time when adding new workers, generated by the code produced by ChatGPT. The output of the code when executed on a local machine is shown on the left side. On the  right side, the same when executed on an available server with 32 CPUs. For the latter, we adjusted the code with \texttt{numWorkers = 1:32} and \texttt{n = 10000}.}
	\label{fig:parallel}
\end{figure}

\PROMPT{In MATLAB, optimize the code for computing that 20 pairs of (theta, theta\_hat) for parallel computing on CPUs. Create a demonstration that shows how the time for performing the same job decreases as the number of workers increases.}
\begin{framed}
	\vspace{-10pt} \begin{verbatim}
% Define the range of theta values
thetas = linspace(0.1, 3, 20);

% Define the number of samples
n = 1000;

% Define the range of workers to use
numWorkers = 1:4;

% Initialize an array to store the execution times
times = zeros(size(numWorkers));

for i = 1:length(numWorkers)
  % Start a timer
  tic

  % Run the loop in parallel using parfor
  parfor (j = 1:length(thetas), numWorkers(i))
    % Generate a sample from the Clayton copula with 
    % parameter theta
    U = ClaytonSample(thetas(j), n);
    % Estimate the parameter theta using maximum likelihood
    ClaytonCopulaMLE(U(:,1), U(:,2));
  end

  % Stop the timer and record the execution time
  times(i) = toc;
end

% Plot the execution times
plot(numWorkers, times, 'o-')
xlabel('Number of workers')
ylabel('Execution time (s)')
	\end{verbatim} \vspace{-10pt}
\end{framed}
We directly got a working solution, which generated the plot shown at the left-hand side of Figure~\ref{fig:parallel}. 
Let us highlight several points:
\begin{itemize}
	\item Being in the context of the conversation, we could afford to be extremely concise and use just one word (``\textsf{that}'' in the first sentence) to point at the functionality we request; natural and time-saving.
	\item We just prompted ChatGPT with a very high-level request, \textsf{Create a demonstration}, and got exactly what we wanted. 
	Originally, we used, say, the traditional way of thinking, which involves being very specific when designing a task. For example, ``create an array for time measurements, measure time for this and that, store it in the array and then plot it in the way that...''.
	Since such tasks probably occur many times in ChatGPT's training data, overly specifying is not necessary, which, again, can save a lot of time. 
	\item Notice the 4 workers in \texttt{numWorkers = 1:4}. It is a common number of CPUs of an average office PC. This number is thus not just a pure random guess, but rather a most occurring one in ChatGPT's training data.
\end{itemize}
Note that apart from our office PC, we executed the code on an available server with 32 CPUs, which generated the plot at the right-hand side of Figure~\ref{fig:parallel}. As only 20 theta values (\texttt{thetas}) are considered, we observe that the improvement in time does not continue from the 21st worker onwards.

\subsubsection{GPUs} \label{sec:gpus}
\PROMPT{In MATLAB, optimize the function ClaytonSample for parallel computing on GPUs. Call the optimized function ClaytonSampleGPUs. Then create a script that measures the run-time of ClaytonSample(2, 1000000) and of ClaytonSampleGPUs(2, 1000000).}
\noindent We got two pieces of code as response. First, ChatGPT provided an optimization of \texttt{ClaytonSample} for GPUs.
\begin{framed}
	\vspace{-10pt} \begin{verbatim}
function [U, V] = ClaytonSampleGPUs(theta, n)
% Generate random variables on the GPU
E1 = gpuArray(single(exprnd(1, n, 1)));
E2 = gpuArray(single(exprnd(1, n, 1)));
V = gpuArray(single(gamrnd(1/theta, 1, n, 1)));
psi = @(t) (1+t).^(-1/theta);
U = gather(psi(E1./V));
V = gather(psi(E2./V));
end
	\end{verbatim} \vspace{-10pt}
\end{framed}
\noindent Then, a code for its demonstration followed.
\begin{framed}
	\vspace{-10pt} \begin{verbatim}
% Define parameters
n = 1000000;
theta = 2;

% Time CPU version
tic
data_cpu = ClaytonSample(theta, n);
time_cpu = toc;

% Time GPU version
tic
data_gpu = ClaytonSampleGPUs(theta, n);
time_gpu = toc;

% Display results
disp(['CPU time: ' num2str(time_cpu) ' seconds']);
disp(['GPU time: ' num2str(time_gpu) ' seconds']);
	\end{verbatim} \vspace{-10pt}
\end{framed}
\noindent After executing the latter, the output was:
\vspace{-10pt} \begin{verbatim}
CPU time: 0.17493 seconds
GPU time: 0.093107 seconds
\end{verbatim} \vspace{-10pt}

As can be observed, the optimization of \texttt{ClaytonSample}, \texttt{ClaytonSampleGPUs}, is based on the addition of \texttt{gpuArray}, which assures that the sampling from the standard exponential and gamma distributions as well as the remaining computations are performed directly on available GPUs. 
The outputs \texttt{U} and \texttt{V} are then gathered from the GPUs onto the client by \texttt{gather}. As the output of the demonstration part of the code shows, this roughly halved the non-optimized run-time.

Apart from a more efficient implementation, ChatGPT also saves our time by exempting us from inspecting whether or not each involved function is supported on GPUs. We should also not forget the educational aspect of the matter. As these optimization techniques are rather advanced, an inexperienced user genuinely learns from these outputs (having in mind they might not always be correct). For example, without explicitly mentioning it in our prompt, the values of \texttt{theta} and \texttt{n} are stored in separate variables before they are used in \texttt{ClaytonSample} and \texttt{ClaytonSampleGPUs}. This belongs to proper coding techniques.

\section{Summary and discussion}\label{sec:summary}
During the development of the working code solving our example task, we observed a considerable list of advantages from which we can benefit while pair programming with ChatGPT. In particular: 
\begin{enumerate}
\item ChatGPT can save time by quickly and concisely summarizing basic facts about the topic of our interest, e.g., formulas or application examples, as illustrated in Section~\ref{sec:warm_up}.
\item  If ChatGPT is not able to provide a correct solution due to a lack of or incorrect knowledge, we can feed it with the correct knowledge, and make it use it to provide a correct solution. In Section~\ref{sec:density}, this approach led ChatGPT to produce a function evaluating the PDF of the copula model in three different programming languages. In Section~\ref{sec:sampling}, a working code for sampling from the copula model is generated once 
ChatGPT was fed by the related non-trivial theory. Particularly the latter example shows that ChatGPT is able to understand even relatively complex concepts, and clearly demonstrates that it can be applied in cases when it faces unknown concepts. 
\item ChatGPT saves time by mapping simple concepts, e.g., sampling from the standard exponential and gamma distributions, to existing code (libraries, APIs, or functions) available for a given programming language, as illustrated in Section~\ref{sec:sampling}. 
\item \label{pros:common} The more common the task to solve, the more successful ChatGPT in generating a correct solution. This is illustrated, e.g., in Section~\ref{sec:estimation}, where we immediately obtained code implementing the maximal likelihood estimator by a simple prompt like \textsf{write code for the maximum likelihood estimator of that parameter}. Another example is the transpilation of the MATLAB solution to Python in Section~\ref{sec:visualization}, or the optimization of existing code for parallel computing on CPUs and GPUs in Section~\ref{sec:parallelization}.
\item ChatGPT can help in cases when an error is thrown after executing the generated code. In Section~\ref{sec:sampling}, we have seen that it not only detected what was wrong, but provided a corrected solution. Apart from saving time needed to search and fix the error, this can be crucial particularly for less experienced programmers, who could find the error too complex and eventually give up. ChatGPT helped us roughly with 1/3 of the errors we encountered. Even if not perfect, this is substantially better than no help at all.
\item ChatGPT can help with creating visualizations. In Section~\ref{sec:visualization}, it generated a visualization suggesting that all previously generated code is correct. Even if we have not asked for it, the visualization included all the typical trivia like labels, benchmarks, limits, legends, etc. 
\item ChatGPT at least partially understands the underlying concepts of what we are doing. Without asking it to do so, it added to the visualization the plot of the identity (see Section~\ref{sec:visualization}), suggesting that it is aware of us trying to estimating the true value of some parameter. 
\item ChatGPT can transpile code from one programming language to another also with high-level prompts like \textsf{Code it in Python} and \textsf{And in R}, demonstrated in Section~\ref{sec:visualization}. The same section also shows that if the transpilation fails (which happened with the transpilation to \textsf{R}), it is possible to use a quick ``brute-force'' solution that also accomplished the task.
\item ChatGPT can optimize the already generated code, e.g., for parallel computations.
By prompting \textsf{optimize that for parallel computing on CPUs}, we immediately got the optimized version of the sample-estimate procedure developed in Section~\ref{sec:visualization}; see Section~\ref{sec:cpus}. The same section also shows that a high-level prompt like \textsf{Create a demonstration of this optimization} can result in code showing the impact of the optimization, again including the typically tedious but necessary trivia like labels, etc. Similarly, such an optimization together with a simple demonstration was generated also for computations on GPUs; see Section~\ref{sec:gpus}. 
\item ChatGPT follows proper coding techniques, so the user can genuinely learn them too. We observed that the produced code is properly commented, indented, modularized, avoids code duplicities, etc. 
\item ChatGPT helps the user to get familiar with the produced code faster.
When providing code, ChatGPT typically surrounds it by further information explaining its main features. To save space, we mostly cut this out, however, an example can be found, e.g., in  Section~\ref{sec:sampling} in connection to the error message thrown by the simple check code. 
\end{enumerate}
We have also seen that pair programming with ChatGPT brings several disadvantages, which should be carefully considered. Let us summarize them and discuss possibilities to mitigate them:
\begin{enumerate}
\item ChatGPT in its current version (early 2023) is poor in reasoning; see Section~\ref{sec:warm_up}. On two examples, we demonstrated how it responses with contradicting answers to the same question. We particularly highlight the case when it first answered \textit{yes} and then \textit{no} to the same question. Also, we demonstrated how dangerous this could be in quantitative reasoning, where it generated incorrect formulas that looked very similar to correct ones; see the PDF derivation in Section~\ref{sec:density}. 
In order to mitigate this problem, a lot of effort can be currently observed. 
One of the most promising examples in the direction of quantitative reasoning is Minerva \citep{lewkowycz2022solving}, an LLM based on the PaLM general language models \citep{chowdhery2022palm} with up to 540 billion of parameters. This model, released in June 2022, gained its attention by scoring 50\% on questions in the MATH data set, which was a significant improvement of the state-of-the-art performance on STEM evaluation datasets; see Table~3 therein. In other works, the authors develop models fine-tuned for understanding mathematical formulas \citep{peng2021mathbert}, or employ deep neural networks in mathematical tasks like symbolic integration or solving differential equations \citep{lample2019deep}. Another way of mitigating the problem can be trying to exploit at maximum the current LLMs by carefully adjusting the prompt in order to get more reliable answers. This increasingly popular technique, called \emph{prompt engineering}, involves special techniques to improve reliability when the model fails on a task\footnote{\href{https://github.com/openai/openai-cookbook/blob/main/techniques\_to\_improve\_reliability.md}{https://github.com/openai/openai-cookbook/blob/main/techniques\_to\_improve\_reliability.md}}, and can substantially improve the solution, e.g., for simple math problems, just by adding ``\textsf{Let's think step by step.}'' at the end of the prompt. Note that we tried this technique in the example considering the tail dependence of the survival Clayton copula in Section~\ref{sec:warm_up}, however, without success, probably because the underlying concepts go beyond simple mathematics.
\item If ChatGPT lacks the necessary knowledge or possesses incorrect knowledge, it may generate an incorrect solution without any indication to the user. As illustrated in Section~\ref{sec:sampling}, 
after asking it for code for sampling from a Clayton copula model, ChatGPT first generated two routines, which were resembling proper sampling algorithms, but were entirely incorrect. Due to the opacity of the current state-of-the-art LLMs that contain tens or even hundreds of billions of parameters, the correctness of the solution can hardly be guaranteed in all cases. 
While there may be efforts to develop more explainable LLMs, it is unlikely that the fundamental challenges related to the complexity of language and the massive amounts of data required for training will be completely overcome.
Therefore, it is essential for a human expert in the field to \emph{always} verify the output generated by the model. 
\item Specifically, ChatGPT tends to be less successful in producing accurate solutions for tasks that are less common. This means that the opposite of  advantage \ref{pros:common}.~also applies. In Section~\ref{sec:density}, this is demonstrated through the probability density function (PDF) of the copula model. In Section~\ref{sec:sampling}, through the sampling algorithm. 
To solve these issues, we provided the required theory to ChatGPT, which led to a correct solution, see the same two sections.
\item ChatGPT does not have any memory. If the conversation is too long, and thus does not fit within ChatGPT's context window, it seems that the model has forgotten some parts of the conversation. This, together with ways how to mitigate this issue, has already been discussed in Section~\ref{sec:interaction}.
\end{enumerate}

Apart from ChatGPT, there are several other language models that are capable of generating code solutions from natural language inputs. One notable example is  AlphaCode \citep{li2022competition}, which achieved on average a ranking of top 54.3\% in competitions with more than 5,000 participants on recent programming competitions on the platform Codeforces. Recently, AlphaCode has been made publicly available\footnote{\href{https://github.com/deepmind/code\_contests}{https://github.com/deepmind/code\_contests}}, including example solutions from the mentioned contest.
Another example is OpenAI Codex\footnote{\href{https://openai.com/blog/openai-codex/}{https://openai.com/blog/openai-codex/}}, already mentioned in the introduction. In contrast to ChatGPT, these models have been developed particularly for code generation. On the one hand, it is thus possible that one can generate solutions that are better than those generated with ChatGPT. It would thus be interesting future research to compare, e.g., the successfulness of these models for solving the tasks considered in this work. 

On the other hand, ChatGPT might be more convenient for many users than these models as it allows for interaction during the coding process. Unlike AlphaCode and OpenAI Codex, which generate code snippets based on natural language inputs without any further interaction, ChatGPT allows users to provide feedback and adjust the generated code in real-time. 
This interaction can be beneficial for several reasons. First, it allows users to clarify their intent and ensures that the generated code aligns with their goals. For example, as we have seen in Section~\ref{sec:sampling} that considers the sampling from a Clayton copula model, if a user requests a specific functionality and the generated code does not quite match what they had in mind, the user can provide feedback to ChatGPT to adjust the code accordingly. Second, the interaction with ChatGPT can help users learn more about programming and improve their coding skills. By engaging in a dialogue with ChatGPT, users can gain insights into the logic and structure of the code they are generating, and learn how to improve their code in the future. For example, in Section~\ref{sec:gpus}, we could genuinely learn how to convert existing code for parallel computing on GPUs. Finally, the interaction with ChatGPT can help users troubleshoot errors and debug their code more effectively. As we have seen in Section~\ref{sec:sampling}, ChatGPT can recognize common programming mistakes, and provide feedback that helps users to identify and fix errors in their code.
These reasons, together with the fact that ChatGPT can be conveniently accessed through a web portal, led us to choose ChatGPT as our pair programming AI partner. 



\section{Conclusion}\label{sec:conclusion}

In a human-AI collaboration, we developed working code that implements sampling from a copula model, estimation of its parameter, visualization suggesting that the last two tasks worked properly, and a parallelization of the code for CPUs as well as for GPUs. To illustrate the coding abilities of the AI part, represented by ChatGPT, all the mentioned tasks were implemented without a single line of code written by the human. In addition to presenting how to achieve a successful solution for a given task, we also showed additional examples demonstrating which modifications of our prompts for ChatGPT turned failed solutions to successful ones. This resulted in a comprehensive list of related pros and cons, suggesting that if typical pitfalls can be avoided, we can substantially benefit from a collaboration with an AI partner like ChatGPT.

\bmhead{Acknowledgments}
The author thanks the Czech Science Foundation (GAČR) for financial support for this work through grant 21-03085S. The author also thanks to Martin Hole\v{n}a and Marius Hofert for constructive comments and recommendations that definitely helped to improve the readability and quality of the paper.

\begin{appendices}

\section{The solution in Python} \label{app:python}

\begin{framed}
	\vspace{-10pt} \begin{verbatim}
import numpy as np
import matplotlib.pyplot as plt
from scipy.optimize import fmin

def ClaytonSample(theta, n):
    E1 = np.random.exponential(size=n)
    E2 = np.random.exponential(size=n)
    V = np.random.gamma(1/theta, 1, size=n)
    psi = lambda t: (1+t)**(-1/theta)
    U = psi(E1/V)
    V = psi(E2/V)
    return np.column_stack((U, V))

def ClaytonCopulaDensity(u, v, theta):
    return ((1+theta)*(u*v)**(-theta-1)*
            (u**(-theta)+v**(-theta)-1)**(-2-1/theta))

def ClaytonCopulaMLE(data):
    n = data.shape[0]
    u, v = data[:, 0], data[:, 1]
    loglik = lambda theta: np.sum(np.log(
             ClaytonCopulaDensity(u, v, theta)))
    thetaMLE = fmin(lambda theta: -loglik(theta), 1, 
                    disp=False)
    return thetaMLE[0]

thetas = np.linspace(0.1, 3, 20)
theta_hats = []
for theta in thetas:
    data = ClaytonSample(theta, 1000)
    theta_hat = ClaytonCopulaMLE(data)
    theta_hats.append(theta_hat)

plt.plot(thetas, theta_hats, 'o')
plt.plot(thetas, thetas, '--')
plt.xlabel('True theta')
plt.ylabel('Estimated theta')
plt.show()
	\end{verbatim} \vspace{-10pt}
\end{framed}
An example of a redundant code is \texttt{n = data.shape[0]}. As can be observed, the variable \texttt{n} has no use in \texttt{ClaytonCopulaMLE}.

\section{The solution in \textsf{R}} \label{app:R}

\begin{framed}
	\vspace{-10pt} \begin{verbatim}
ClaytonSample <- function(theta, n) {
	E1 <- rexp(n)
	E2 <- rexp(n)
	V <- rgamma(n, 1/theta, 1)
	psi <- function(t) (1+t)^(-1/theta)
	U <- psi(E1/V)
	V <- psi(E2/V)
	return(cbind(U, V))
}

ClaytonCopulaDensity <- function(u, v, theta) {
	return((1+theta)*(u*v)^(-theta-1)*
	       (u^(-theta)+v^(-theta)-1)^(-2-1/theta))
}

ClaytonCopulaMLE <- function(data) {
	u <- data[,1]
	v <- data[,2]
	loglik <- function(theta) -sum(log(
	           ClaytonCopulaDensity(u, v, theta)))
	thetaMLE <- optimize(loglik, c(0.1, 3), maximum=FALSE)$minimum
	return(thetaMLE)
}

thetas <- seq(0.1, 3, length.out=20)
theta_hats <- c()
for (theta in thetas) {
	data <- ClaytonSample(theta, 1000)
	theta_hat <- ClaytonCopulaMLE(data)
	theta_hats <- c(theta_hats, theta_hat)
}

plot(thetas, theta_hats, pch=19, col="blue", ylim=c(0,3),
xlab="True theta", ylab="Estimated theta")
abline(a=0, b=1, lty=2)
	\end{verbatim} \vspace{-10pt}
\end{framed}
Interestingly, even if this code is a direct transpilation of the code from Appendix~\ref{app:python}, the redundant code from the Python version of \texttt{ClaytonCopulaMLE} is not present. This hints on the ability of ChatGPT to keep only the code that is relevant.

\end{appendices}

%% file: paper_arxiv.bbl
\begin{thebibliography}{}

\bibitem[Bang et~al., 2023]{bang2023multitask}
Bang, Y., Cahyawijaya, S., Lee, N., Dai, W., Su, D., Wilie, B., Lovenia, H.,
  Ji, Z., Yu, T., Chung, W., et~al. (2023).
\newblock A multitask, multilingual, multimodal evaluation of chatgpt on
  reasoning, hallucination, and interactivity.
\newblock {\em arXiv preprint arXiv:2302.04023}.

\bibitem[Boulin, 2022]{boulin2022sample}
Boulin, A. (2022).
\newblock Sample from copula: a coppy module.
\newblock {\em arXiv preprint arXiv:2203.17177}.

\bibitem[Brown et~al., 2020]{brown2020language}
Brown, T., Mann, B., Ryder, N., Subbiah, M., Kaplan, J.~D., Dhariwal, P.,
  Neelakantan, A., Shyam, P., Sastry, G., Askell, A., et~al. (2020).
\newblock Language models are few-shot learners.
\newblock {\em Advances in neural information processing systems},
  33:1877--1901.

\bibitem[Chen et~al., 2021]{chen2021evaluating}
Chen, M., Tworek, J., Jun, H., Yuan, Q., Pinto, H. P. d.~O., Kaplan, J.,
  Edwards, H., Burda, Y., Joseph, N., Brockman, G., et~al. (2021).
\newblock Evaluating large language models trained on code.
\newblock {\em arXiv preprint arXiv:2107.03374}.

\bibitem[Chowdhery et~al., 2022]{chowdhery2022palm}
Chowdhery, A., Narang, S., Devlin, J., Bosma, M., Mishra, G., Roberts, A.,
  Barham, P., Chung, H.~W., Sutton, C., Gehrmann, S., et~al. (2022).
\newblock Palm: Scaling language modeling with pathways.
\newblock {\em arXiv preprint arXiv:2204.02311}.

\bibitem[Christiano et~al., 2017]{christiano2017deep}
Christiano, P.~F., Leike, J., Brown, T., Martic, M., Legg, S., and Amodei, D.
  (2017).
\newblock Deep reinforcement learning from human preferences.
\newblock {\em Advances in neural information processing systems}, 30.

\bibitem[Clayton, 1978]{clayton_1978}
Clayton, D.~G. (1978).
\newblock A model for association in bivariate life tables and its application
  in epidemiological studies of familial tendency in chronic disease incidence.
\newblock {\em Biometrika}, 65:141--151.

\bibitem[Devlin et~al., 2018]{devlin2018bert}
Devlin, J., Chang, M.-W., Lee, K., and Toutanova, K. (2018).
\newblock {BERT}: Pre-training of deep bidirectional transformers for language
  understanding.
\newblock {\em arXiv preprint arXiv:1810.04805}.

\bibitem[Frieder et~al., 2023]{frieder2023mathematical}
Frieder, S., Pinchetti, L., Griffiths, R.-R., Salvatori, T., Lukasiewicz, T.,
  Petersen, P.~C., Chevalier, A., and Berner, J. (2023).
\newblock Mathematical capabilities of chatgpt.
\newblock {\em arXiv preprint arXiv:2301.13867}.

\bibitem[Hob{\ae}k~Haff, 2013]{hobaek2013parameter}
Hob{\ae}k~Haff, I. (2013).
\newblock Parameter estimation for pair-copula constructions.

\bibitem[Hofert, 2010]{Hof10book}
Hofert, M. (2010).
\newblock Sampling nested {A}rchimedean copulas with applications to {CDO}
  pricing.

\bibitem[Hofert et~al., 2018]{hofert2018hierarchical}
Hofert, M., Huser, R., and Prasad, A. (2018).
\newblock Hierarchical {A}rchimax copulas.
\newblock {\em Journal of Multivariate Analysis}, 167:195--211.

\bibitem[Hofert et~al., 2013]{Hofert13}
Hofert, M., M{\"a}chler, M., and McNeil, A.~J. (2013).
\newblock Archimedean copulas in high dimensions: Estimators and numerical
  challenges motivated by financial applications.
\newblock {\em Journal de la Soci{\'e}t{\'e} Fran{\c{c}}aise de Statistique},
  154(1):25--63.

\bibitem[Huang et~al., 2022]{huang2022spatio}
Huang, Y., Zhang, B., Pang, H., Wang, B., Lee, K.~Y., Xie, J., and Jin, Y.
  (2022).
\newblock Spatio-temporal wind speed prediction based on clayton copula
  function with deep learning fusion.
\newblock {\em Renewable Energy}, 192:526--536.

\bibitem[Joe, 2014]{joe2014dependence}
Joe, H. (2014).
\newblock {\em Dependence Modeling with Copulas}.
\newblock CRC Press.

\bibitem[Katz et~al., 2023]{katz2023gpt}
Katz, D.~M., Bommarito, M.~J., Gao, S., and Arredondo, P. (2023).
\newblock {GPT}-4 passes the bar exam.
\newblock {\em Available at SSRN 4389233}.

\bibitem[Lample and Charton, 2019]{lample2019deep}
Lample, G. and Charton, F. (2019).
\newblock Deep learning for symbolic mathematics.
\newblock {\em arXiv preprint arXiv:1912.01412}.

\bibitem[Lewkowycz et~al., 2022]{lewkowycz2022solving}
Lewkowycz, A., Andreassen, A., Dohan, D., Dyer, E., Michalewski, H., Ramasesh,
  V., Slone, A., Anil, C., Schlag, I., Gutman-Solo, T., et~al. (2022).
\newblock Solving quantitative reasoning problems with language models.
\newblock {\em arXiv preprint arXiv:2206.14858}.

\bibitem[Li et~al., 2022]{li2022competition}
Li, Y., Choi, D., Chung, J., Kushman, N., Schrittwieser, J., Leblond, R.,
  Eccles, T., Keeling, J., Gimeno, F., Dal~Lago, A., et~al. (2022).
\newblock Competition-level code generation with alphacode.
\newblock {\em Science}, 378(6624):1092--1097.

\bibitem[Liu et~al., 2019]{liu2019roberta}
Liu, Y., Ott, M., Goyal, N., Du, J., Joshi, M., Chen, D., Levy, O., Lewis, M.,
  Zettlemoyer, L., and Stoyanov, V. (2019).
\newblock Ro{BERT}a: A robustly optimized bert pretraining approach.
\newblock {\em arXiv preprint arXiv:1907.11692}.

\bibitem[Maddigan and Susnjak, 2023]{maddigan2023chat2vis}
Maddigan, P. and Susnjak, T. (2023).
\newblock Chat2vis: Generating data visualisations via natural language using
  chatgpt, codex and gpt-3 large language models.
\newblock {\em arXiv preprint arXiv:2302.02094}.

\bibitem[Marshall and Olkin, 1988]{marshall1988families}
Marshall, A.~W. and Olkin, I. (1988).
\newblock Families of multivariate distributions.
\newblock {\em Journal of the American Statistical Association},
  83(403):834--841.

\bibitem[McNeil et~al., 2015]{mcneil2015quantitative}
McNeil, A., Frey, R., and Embrechts, P. (2015).
\newblock {\em Quantitative risk management: Concepts, techniques and tools}.
\newblock Princeton university press.

\bibitem[Michimae and Emura, 2022]{michimae2022likelihood}
Michimae, H. and Emura, T. (2022).
\newblock Likelihood inference for copula models based on left-truncated and
  competing risks data from field studies.
\newblock {\em Mathematics}, 10(13):2163.

\bibitem[Nelsen, 2006]{Nel06}
Nelsen, R.~B. (2006).
\newblock {\em An Introduction to Copulas}.
\newblock Springer-Verlag, 2nd edition.

\bibitem[Open{AI}, 2023]{gpt4}
Open{AI} (2023).
\newblock {GPT}-4 technical report.
\newblock {\em arXiv preprint arXiv:2303.08774}.

\bibitem[Ouyang et~al., 2022]{ouyang2022training}
Ouyang, L., Wu, J., Jiang, X., Almeida, D., Wainwright, C.~L., Mishkin, P.,
  Zhang, C., Agarwal, S., Slama, K., Ray, A., et~al. (2022).
\newblock Training language models to follow instructions with human feedback.
\newblock {\em arXiv preprint arXiv:2203.02155}.

\bibitem[Peng et~al., 2021]{peng2021mathbert}
Peng, S., Yuan, K., Gao, L., and Tang, Z. (2021).
\newblock Mathbert: A pre-trained model for mathematical formula understanding.
\newblock {\em arXiv preprint arXiv:2105.00377}.

\bibitem[Schellhase and Spanhel, 2018]{schellhase2018estimating}
Schellhase, C. and Spanhel, F. (2018).
\newblock Estimating non-simplified vine copulas using penalized splines.
\newblock {\em Statistics and Computing}, 28:387--409.

\bibitem[Sklar, 1959]{Skl59}
Sklar, A. (1959).
\newblock Fonctions de r\'{e}partition a n dimensions et leurs marges.
\newblock {\em Publications de l’Institut Statistique de l’Université de
  Paris}, 8:229--231.

\bibitem[Stiennon et~al., 2020]{stiennon2020learning}
Stiennon, N., Ouyang, L., Wu, J., Ziegler, D., Lowe, R., Voss, C., Radford, A.,
  Amodei, D., and Christiano, P.~F. (2020).
\newblock Learning to summarize with human feedback.
\newblock {\em Advances in Neural Information Processing Systems},
  33:3008--3021.

\bibitem[Williams, 2001]{williams2001integrating}
Williams, L. (2001).
\newblock Integrating pair programming into a software development process.
\newblock In {\em Proceedings 14th Conference on Software Engineering Education
  and Training. '{I}n search of a software engineering profession' ({C}at.
  {N}o. {PR}01059)}, pages 27--36. IEEE.

\end{thebibliography}
